\documentclass[runningheads]{llncs}

\usepackage{eccv}

\usepackage{eccvabbrv}

\usepackage{graphicx}
\usepackage{booktabs}
\usepackage{wrapfig} 
\usepackage[table]{xcolor}
\usepackage{xcolor}
\usepackage{multirow}
\definecolor{codegray}{gray}{0.95}

\usepackage[accsupp]{axessibility}  

\usepackage{hyperref}

\usepackage{orcidlink}

\begin{document}

\title{Targeted Structure Completion for Sparse-View 3D Reconstruction in Autonomous Driving}

\titlerunning{FocusGS}

\author{Guoqing Wang\inst{1} 
\and Pin Tang\inst{1} 
\and Xiangxuan Ren\inst{1} 
\and Liping Hou\inst{2} 
\and Chao Ma\inst{1}, \thanks{Corresponding author.}
}

\authorrunning{G.~Wang et al.}

\institute{$^1$ MoE Key Lab of Artificial Intelligence, Institute of AI, \\
Shanghai Jiao Tong University, Shanghai, China \\
$^2$ Central Research Institute, Huawei, Beijing, China
{\tt\small \{guoqing.wang,pin.tang,bunny\_renxiangxuan,chaoma\}@sjtu.edu.cn} \\
{\tt\small houliping1@huawei.com}\\
{\small Project page: \url{https://focusgs.github.io/}}
}

\maketitle

\begin{abstract}
Reconstructing 3D scene structures from sparse, low-overlap observations remains a fundamental challenge in autonomous driving. Recent state-of-the-art frameworks achieve promising results by incorporating voxel-based Gaussians, but incur substantial computational redundancy due to a uniform volumetric processing strategy. To bridge the gap between the efficiency of pixel-based Gaussian methods and the structural completeness of voxel-based Gaussian approaches, we propose \textbf{FocusGS}, a simple yet effective framework that shifts the paradigm from global densification to targeted structural completion. Our central insight is that structural completion should be decoupled from deterministic regions, with computation concentrated exclusively on areas exhibiting geometric ambiguity. Specifically, FocusGS addresses the localization challenge by deriving a 3D Geometric Ambiguity Manifold to accurately isolate localized areas prone to occlusion and high geometric uncertainty. To overcome the subsequent manifold completion challenge, we design a lightweight targeted structure completion module that selectively instantiates and optimizes continuous Gaussian queries strictly within this unstructured, sparse topological subspace. Extensive experiments demonstrate that FocusGS achieves a superior efficiency-quality trade-off, advancing state-of-the-art performance on driving-centric benchmarks while naturally reducing the total number of Gaussians by $\sim$74\% and decreasing rendering time by $\sim$34\%.
  \keywords{3D Scene Reconstruction \and Geometric Ambiguity \and Targeted Structure Completion}
\end{abstract}

\section{Introduction}
\label{sec:intro}
Reconstructing 3D scene structures from sparse observations is a fundamental problem in computer vision and graphics~\cite{zhou2024drivinggaussian, song2025adgaussian, wei2025omni, chen2024mvsplat, charatan2024pixelsplat}. Recent feed-forward frameworks~\cite{chen2024mvsplat, charatan2024pixelsplat}, built upon 3D Gaussian Splatting (3DGS)~\cite{kerbl20233d}, directly predict pixel-based 3D Gaussians from sparse inputs in a single forward pass. These approaches offer remarkable efficiency and generalization, particularly in large-scale dynamic environments. However, they implicitly assume substantial overlap between input views to enforce geometric consistency. This assumption confronts the challenge in ego-centric autonomous driving, where cameras are designed for maximal coverage with minimal overlap ($<15\%$), leading to degraded reconstruction quality.

\begin{figure*}[!t]
\centering
\includegraphics[width=1.0\textwidth]{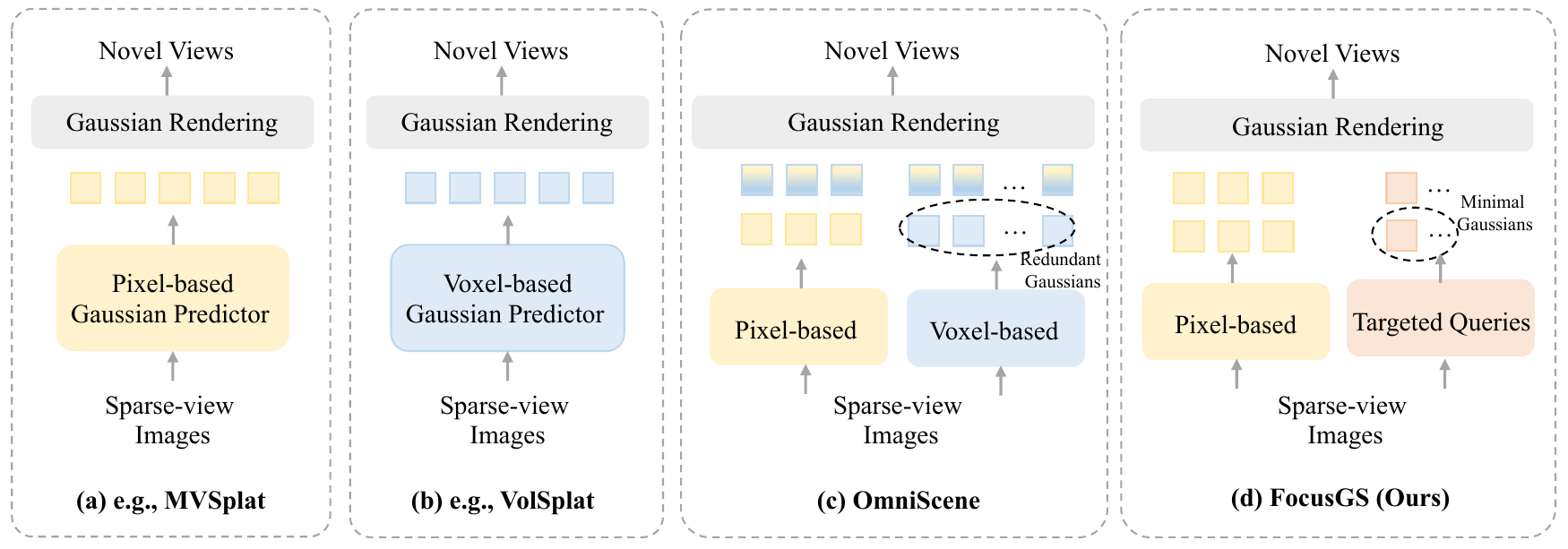}
\caption{
Comparison of feed-forward 3D Gaussian Splatting methods. (a, b) Prior works rely on either pixel-~\cite{charatan2024pixelsplat, chen2024mvsplat} or voxel-based~\cite{wang2025volsplat} predictors. (c) Omni-Scene~\cite{wei2025omni} combines both but suffers from redundant Gaussians. (d) Our FocusGS introduces targeted structure completion based on pixel-based features, achieving high-fidelity reconstruction with minimal Gaussian overhead.
}
\label{fig:intro}
\end{figure*}

To alleviate the reliance on cross-view overlap, voxel-based Gaussian methods~\cite{huang2024gaussianformer, zhu2025voxelsplat, wang2025volsplat} lift 2D image features into 3D space, thereby improving robustness to occlusion and frustum truncation (Fig.~\ref{fig:intro}(b)). In particular, state-of-the-art Omni-Gaussian frameworks~\cite{wei2025omni} adopt a dual-branch architecture that integrates pixel- and voxel-based Gaussians to infer and complete occluded regions explicitly. Although effective, this design introduces substantial computational overhead due to its uniform volumetric processing strategy (Fig.~\ref{fig:intro}(c)), which instantiates and optimizes millions of Gaussians throughout the entire 3D space. However, in typical driving scenarios, most regions (e.g., flat roads or clear skies) are geometrically continuous and largely free of occlusion, and can therefore be adequately represented by pixel-based Gaussians alone. As a result, the volumetric branch redundantly processes extensive simple surfaces, effectively allocating a large number of voxel-based Gaussians to rectify geometric defects that are confined to only a small subset of the scene.

\begin{wrapfigure}{r}{6.2cm}
       \includegraphics[width=6.2cm]{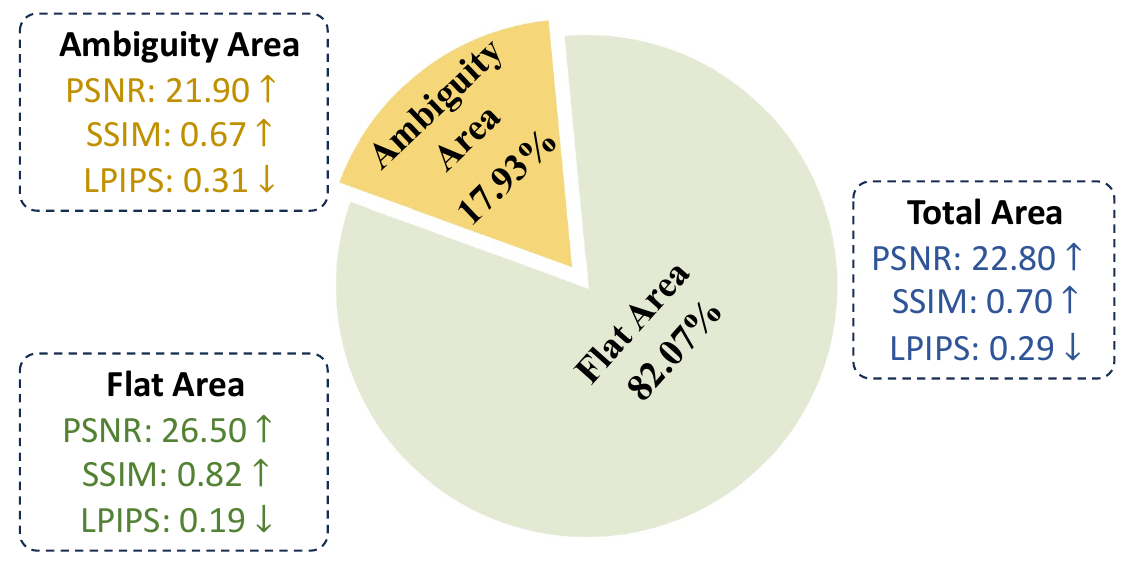}
        \caption{Quantitative analysis of pixel-based methods across different geometric areas.}
        \label{fig:com}
\end{wrapfigure}

To quantitatively illustrate this inefficiency, we analyzed the performance of pixel-based methods across different geometric regions. As illustrated in Fig.~\ref{fig:com}, a vast majority of the environment consists of flat, deterministic areas where base representations already achieve high visual fidelity. Conversely, severe degradation occurs in a small fraction of the scene characterized by geometric ambiguity. This stark performance disparity motivates a key question:
\textit{Can we achieve the efficiency of pixel-based Gaussian methods while preserving the structural completeness as Omni-Gaussian does?} We argue that targeted structural
completion in critical regions provides a simple and principled solution. Rather than uniformly generating Gaussians across the entire 3D volume, the model should
selectively concentrate on critical areas. We formally define these critical areas as \textit{geometric ambiguity manifold}, i.e., a localized, sparse 3D topological subspace encompassing regions with high geometric uncertainty, such as severe depth discontinuities and occlusion boundaries. However, conducting structural completion strictly on this manifold presents two fundamental challenges. First, the localization challenge involves precisely identifying and isolating these occlusion-prone regions, which is non-trivial but essential for minimizing the number of redundant Gaussians. Second, since the ambiguity manifold is spatially irregular and sparsely distributed, traditional dense voxel grids become computationally inefficient. The targeted structure completion should therefore operate on unstructured and continuous regions without relying on dense contextual neighborhoods. Moreover, it must remain lightweight and seamlessly integrate with the base representation, avoiding the substantial overhead introduced by full voxel-based Gaussians.

To this end, we propose \textbf{FocusGS}, a simple yet effective framework for sparse-view 3D reconstruction in autonomous driving shown in Fig.~\ref{fig:intro}(d). Our key insight is that geometric ambiguity in ego-centric scenes is inherently non-uniform and primarily concentrated near visibility transitions. To address the localization challenge, FocusGS first constructs pixel-based Gaussians as the base representation and subsequently derives the 3D geometric ambiguity manifold by lifting 2D ambiguity regions into a sparse 3D uncertainty subspace. This manifold serves as a precise spatial prior for incomplete geometry. To tackle the manifold completion challenge, we design a lightweight targeted structure completion module that randomly initializes a set of continuous queries strictly within this 3D uncertainty subspace. Rather than operating over the entire volume, the model selectively instantiates and optimizes targeted Gaussian queries only on the manifold. This strategy effectively decouples structural completion from deterministic regions, enabling FocusGS to resolve geometric ambiguities with a fraction of the Gaussians required by Omni-Scene.

Experiments on multiple benchmarks validate the effectiveness of FocusGS. Compared with Omni-Scene~\cite{wei2025omni}, FocusGS improves the efficiency-quality trade-off, reducing the total number of Gaussians by approximately \textbf{74\%} and rendering time by roughly \textbf{34\%}. Our main contributions are summarized as follows:
\begin{itemize} 
\item We identify uniform volumetric processing as a critical bottleneck in existing dual-branch pipelines and leverage quantitative area analysis to motivate a novel targeted structure completion paradigm.
\item We introduce FocusGS, which instantiates this paradigm by updating 3D Gaussians on the geometric ambiguity manifold by initializing targeted queries within an irregular, sparse subspace. This design allows for precise structural completion without incurring the cost of dense volumetric processing.
\item Extensive experiments demonstrate that FocusGS achieves state-of-the-art reconstruction quality on nuScenes while naturally reducing redundant Gaussians by approximately 74\% and rendering time by approximately 34\%.
\end{itemize}

\section{Related Work}

\subsection{3D Scene Reconstruction}
Previous 3D scene reconstruction methods~\cite{yu2021pixelnerf, wang2021ibrnet, liu2022neural, chen2021mvsnerf, johari2022geonerf, wu2024recent, cao2023hexplane, chen2024pgsr} incorporate structural priors to predict implicit neural fields~\cite{mildenhall2020nerf}, light fields~\cite{suhail2022light}, or explicit 3D Gaussians~\cite{kerbl20233d} in a single forward pass. While Neural Radiance Fields (NeRF)~\cite{mildenhall2020nerf} achieve high fidelity, their reliance on dense per-ray sampling imposes significant computational overhead, a bottleneck that persists even in generalizable feed-forward variants~\cite{yu2021pixelnerf,wang2021ibrnet,chen2021mvsnerf}. Although light-field approaches improve efficiency, they inherently sacrifice explicit 3D geometry~\cite{mildenhall2019local,sitzmann2021light}. Consequently, explicit representations like 3D Gaussian Splatting (3DGS)~\cite{kerbl20233d} have emerged, utilizing differentiable rasterization to achieve real-time rendering. Building on this, recent feed-forward frameworks~\cite{chen2021mvsnerf,charatan2024pixelsplat, xu2025depthsplat, jiang2025anysplat, shi2025unisplat} directly predict pixel-aligned Gaussians from sparse views using geometric priors. However, these methods heavily rely on substantial cross-view overlap to enforce geometric consistency. In ego-centric autonomous driving, where cameras are designed to maximize spatial coverage, cross-view overlap is typically minimal (< 15\%). This causes traditional pixel-based methods to fail under severe occlusion and frustum truncation, significantly degrading reconstruction quality. Moreover, applying dense or uniform completion over the entire scene introduces substantial redundancy, as most well-observed regions already contain sufficient geometric evidence. To address these limitations, our work proposes a novel targeted structure completion paradigm to effectively resolve geometric ambiguities in sparse-view environments.

\subsection{Gaussian Splatting in Autonomous Driving}
The adaptation of 3D Gaussian Splatting (3DGS)~\cite{kerbl20233d, yu2024mip, ren2024octree} has driven numerous autonomous driving applications~\cite{ren2024l4gm, wang2024occgen, tang2024sparseocc, tian2024occ3d, wei2024occllama, zhang2023occformer, huang2021bevdet, li2022bevformer, li2023bevdepth, liu2023bevfusion, hu2023planning, jiang2023vad, wei2023surroundocc, wang2023openoccupancy, jia2023adriver, cao2022monoscene, chen2024ppad, zhou2025opendrivevla, zhou2025autovla, lu2025wovogen, min2024driveworld, wang2024drivedreamer, wang2024worlddreamer, lu2024drivingrecon, jiang2025gausstr}, particularly in scene reconstruction~\cite{zhou2024drivinggaussian, lu2024drivingrecon, song2025adgaussian, huang2024gaussianformer, yan2024street}. Previous works have predominantly focused on per-scene optimization, achieving high-fidelity reconstructions by exploiting comprehensive sensor data from specific sequences. For instance, StreetGaussians~\cite{yan2024street} models dynamic urban environments by separating static backgrounds and moving vehicles into distinct Gaussian sets, employing a layered optimization strategy. Similarly, Gaussian methods for occupancy~\cite{huang2024gaussianformer, gan2025gaussianocc, zuo2025gaussianworld} extend scene encoding with semantic Gaussians, where each primitive serves as a flexible region of interest encapsulating both geometric and semantic features. Recent generalizable models~\cite{chen2024mvsplat, charatan2024pixelsplat} circumvent this cost by using feed-forward networks to predict Gaussians directly from sparse views. Despite these advancements, ego-centric driving reconstruction remains inherently challenged by minimal cross-view overlap and frequent occlusions. To handle this, Omni-Scene~\cite{wei2025omni} introduces a dual-branch representation combining pixel- and voxel-based Gaussians. However, its uniform volumetric processing generates Gaussians across the entire 3D space indiscriminately, leading to massive computational overhead. Our FocusGS resolves this bottleneck by shifting the paradigm from global densification to targeted structural completion, efficiently targeting geometric ambiguities to preserve high-fidelity accuracy while substantially reducing the total Gaussian count.

\begin{figure*}[!t]
\centering
\includegraphics[width=1.0\textwidth]{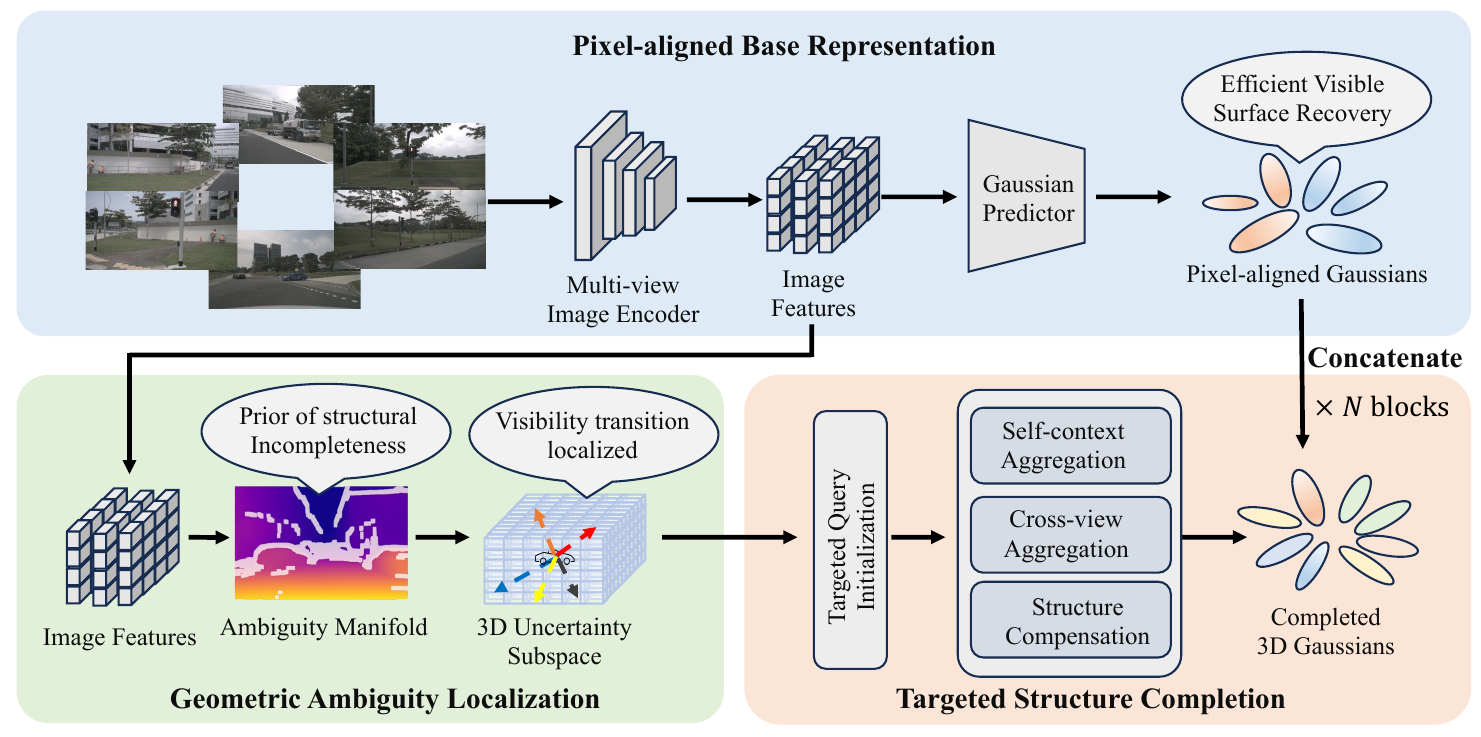}
\caption{The overall pipeline of FocusGS. Given sparse multi-view images, we first generate a pixel-aligned base Gaussian representation for continuous visible surfaces. To handle occlusions and depth discontinuities, we explicitly localize regions of high uncertainty by constructing a geometric ambiguity manifold. A targeted structure completion module then initializes and optimizes Gaussian queries exclusively within this derived 3D uncertainty subspace, naturally bridging the gap between pixel-based efficiency and volumetric structural completeness.}
\label{fig:overview}
\end{figure*}

\section{Proposed Method}

\subsection{Overall Architecture}
FocusGS introduces a \textbf{Targeted Structure Completion} paradigm to infer 3D Gaussians from sparse viewpoints as illustrated in Fig.~\ref{fig:overview}. Instead of indiscriminately processing the entire 3D volume, our framework spatially decouples deterministic areas from geometrically uncertain regions, applying structural completion exclusively where it is needed. Given sparse multi-view RGB inputs $\{I_i\}_{i=1}^{N} \in \mathbb{R}^{N \times H \times W \times 3}$, we first extract $4\times$ down-sampled image features $\{F_i\}_{i=1}^N \in \mathbb{R}^{N \times \frac{H}{4} \times \frac{W}{4} \times C}$ using a 2D pretrained image backbone. We formulate the base generation as a mapping function $\mathcal{M}$ that transforms these down-sampled features into a set of pixel-aligned 3D Gaussians:
\begin{equation}\mathcal{M}:\{F_i\}_{i=1}^{N}\rightarrow \{(\boldsymbol{\delta}_j,\ \boldsymbol{\alpha}_j,\ \boldsymbol{s}_j,\ \boldsymbol{q}_j,\ \boldsymbol{c}_j)\}_{j=1}^{K},
\end{equation}
where $K$ denotes the total number of 3D Gaussians. The variables $\boldsymbol{\delta}_j$, $\boldsymbol{\alpha}_j$, $\boldsymbol{s}_j$, $\boldsymbol{q}_j$, and $\boldsymbol{c}_j$ represent the learned spatial offset, opacity, scale, rotation quaternion, and RGB color, respectively.

To construct this base pixel-aligned representation, the extracted image features are augmented with geometric and view-specific priors, such as Plücker ray encodings, camera embeddings, and pseudo-depth, to facilitate robust multi-view context aggregation. These geometrically enriched features are then projected to predict per-pixel depth and the corresponding 3D Gaussian attributes. 
To determine the precise center $\boldsymbol{\mu}_p$ of each
Gaussian, we unproject the pixel from the ray origin $\mathbf{o}_p$ along the ray direction $\boldsymbol{r}_p$ using the predicted depth $\boldsymbol{d}_p$. This coarse position is subsequently refined using the learned spatial offset $\boldsymbol{\delta}_p \in \mathbb{R}^3$, formulated as:
\begin{equation}
\boldsymbol{\mu}_p = \boldsymbol{o}_p + \boldsymbol{d}_p\boldsymbol{r}_p + \boldsymbol{\delta}_p.
\end{equation}

Through these steps, we obtain the base pixel-aligned Gaussians. While this base representation efficiently models deterministic regions (e.g., flat roads and clear skies), it inevitably struggles to recover complete geometry in 3D space heavily affected by occlusion and visibility transitions. To address this limitation without resorting to the computationally prohibitive uniform volumetric densification seen in prior dual-branch methods (e.g., Omni-Scene~\cite{wei2025omni}), FocusGS introduces a spatially decoupled architecture. First, an ambiguity localization module extracts a 2D \textit{geometric ambiguity manifold} from the aggregated features to explicitly identify regions requiring structural restoration. This precise spatial prior is then lifted into a sparse 3D uncertainty subspace. Finally, a lightweight targeted structural completion module selectively instantiates and optimizes additional Gaussian queries strictly within this restricted volume. This targeted design ensures high-fidelity reconstruction at complex occlusions while drastically reducing computational redundancy.

\subsection{Geometric Ambiguity Localization}
\label{sec:location}
\begin{figure*}[!t]
\centering
\includegraphics[width=1.0\textwidth]{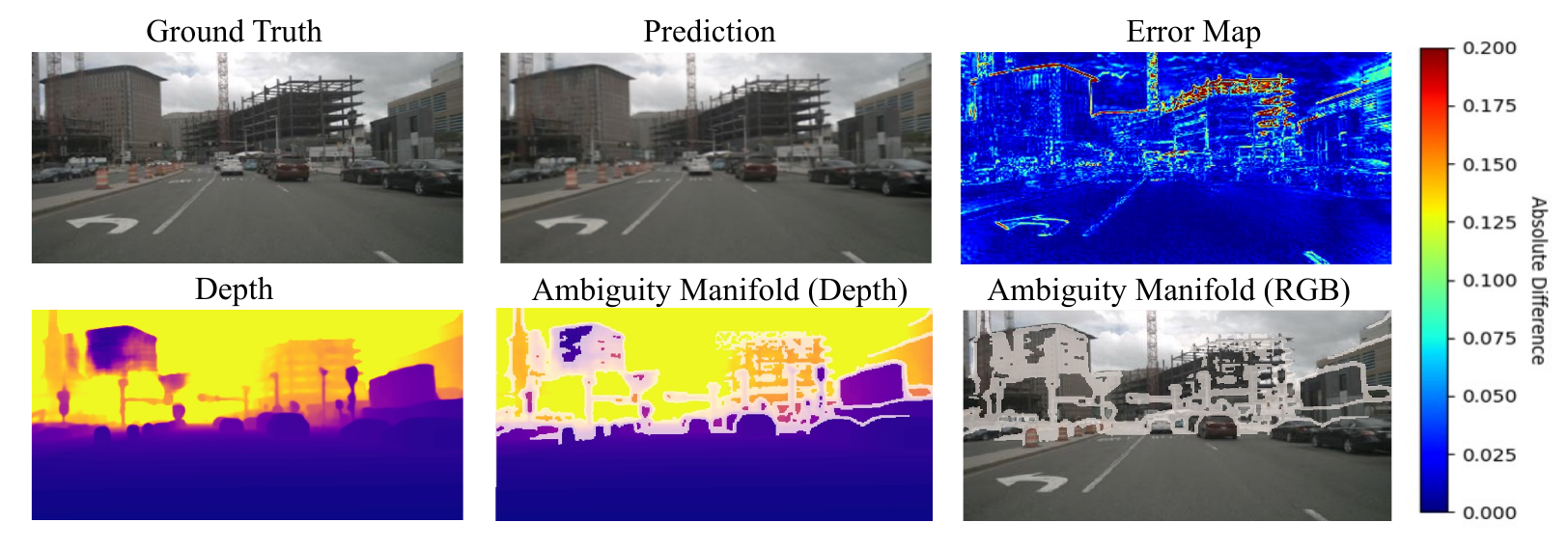}
\caption{Visualizing the correlation between reconstruction error and geometric ambiguity manifold. Structural artifacts mainly occur at depth boundaries, and our extracted 2D geometric ambiguity prior aligns closely with these high-error regions, effectively guiding updates to geometrically ambiguous areas.}
\label{fig:pre}
\end{figure*}

While pixel-based unprojection efficiently and accurately handles the vast majority of deterministic surfaces, severe reconstruction errors predominantly arise at visibility transitions, where sightlines shift abruptly between foreground and background. As demonstrated in Fig.~\ref{fig:com}, pixel-based methods achieve satisfactory performance in geometrically simple regions but suffer significant degradation in ambiguous areas. Furthermore, our empirical analysis reveals that these reconstruction errors are strongly correlated with depth discontinuities; as illustrated in Fig.~\ref{fig:pre}, the spatial distribution of the error map highly overlaps with our extracted 2D prior. To address the \textit{localization challenge}, we explicitly detect these visibility transitions to form a 2D \textit{geometric ambiguity manifold}, which is subsequently lifted into a sparse 3D uncertainty subspace for targeted structural completion. 

\subsubsection{2D Geometric Ambiguity Manifold} In surround-view images, the depth at each pixel represents the distance to the first visible surface along its camera ray. Within continuous surfaces, depth varies smoothly. However, at an occlusion boundary, the visible surface abruptly transitions from the foreground to the background, inducing a sharp depth discontinuity. We exploit the large depth gradients yielded by finite differences at these switches to effectively localize geometric ambiguity. Specifically, given the predicted depth map $Z_i$ for each multi-view image, we compute central-difference gradients as:
\begin{equation}
    D_x=Z_i \times K_x,\ D_y=Z_i \times K_y,\ E_i=\sqrt{D_x^{\,2}+D_y^{\,2}},
\end{equation}
where $K_x=[-1,0,1]$, $K_y=[-1,0,1]^\top$, and the operator $\times$ denotes 2D convolution.
We further introduce a hyperparameter $\tau_g$ to threshold the computed gradient magnitudes, isolating the strict boundaries between foreground and background:
\begin{equation}
    \mathcal{O}_{b}(x, y) =
\begin{cases}
1, & \text{if } \! \ E_{i}(x, y) \geq \tau_g \\
0, & \text{otherwise}
\end{cases}
\end{equation}
where $\{x, y\}$ denote the corresponding spatial coordinates on the image plane. Because geometric uncertainty typically spans a neighborhood around the exact boundary, we apply morphological dilation with a square structuring element. This expands the strict boundaries into continuous uncertainty bands, yielding the 2D geometric ambiguity manifold $\mathcal{O}_d$ that robustly localizes occlusion-prone or geometry-discontinuous regions. Generally, for a square structuring window of size $k$ (with radius $r=\frac{k-1}{2}$), we obtain the dilated manifold $\mathcal{O}_{d}$ as:
\begin{equation}
\mathcal{O}_d(x,y) \;=\; \mathbf{1}\!\left( \sum_{i=-r}^{r}\sum_{j=-r}^{r} \mathcal{O}_b(x+i,\,y+j) \;>\; 0 \right),
\label{eq:dilation}
\end{equation}
where $\mathbf{1}(\cdot)$ is the indicator function. 
This dilation sets \(\mathcal{O}_d(x,y)\) to 1 if at least one boundary pixel in 
\(\mathcal{O}_b\) falls within the \(k\times k\) neighborhood centered at \((x,y)\).

\subsubsection{3D Uncertainty Subspace Lifting} 
Given the per-view 2D ambiguity manifolds $\mathcal{O}_d\in\{0,1\}^{H\times W}$, we lift them into a sparse 3D neighborhood around the depth discontinuities, forming what we define as the \textit{3D uncertainty subspace} $\mathcal{O}_{3d} \subset\mathbb R^3$. We project the ambiguous pixels into camera rays and subsequently transform them into the world coordinate system. For each localized pixel, we take its predicted depth $d_p$ as the midpoint of a 3D line segment and compute the near and far endpoints, $[\mathbf{p_0}, \mathbf{p_1}]$, as:
\begin{equation}
\begin{split}
    \mathbf{p_0} = \mathbf{o}_p + (d_{p} - \Delta_p)\mathbf{r}_p,\\
    \mathbf{p_1} = \mathbf{o}_p + (d_{p} + \Delta_p)\mathbf{r}_p.
\end{split}
\end{equation}
Here, $\Delta_p$ denotes the longitudinal thickness along the ray, which dynamically adapts to depth scale and uncertainty. It is expressed as:
\begin{equation}
    \Delta_p = \kappa_{\mathrm{rel}} d_p + \kappa_{\mathrm{abs}} .
\end{equation}
where $\kappa_{\mathrm{rel}}$ and $\kappa_{\mathrm{abs}}$ are hyperparameters controlling the allowed longitudinal variance. This formulation ensures the thickness increases linearly with depth, granting greater longitudinal uncertainty for farther points. We posit that sampling along the segment $[\mathbf{p_0}, \mathbf{p_1}]$ comprehensively covers the 3D spatial distribution of the occluded geometry. The final 3D uncertainty subspace $\mathcal{O}_{3d}$ is constructed as the union of all lifted segments across the input views:
\begin{equation}
    \mathcal{O}_{3d}=\bigcup_{v\in\mathcal V}\ \bigcup_{\mathcal{O}_d^v(x,y)=1}\ \mathcal S_{x,y}^{v},
\end{equation}
where $S_{x,y}^{v}$ denotes the 3D segment originating from position $\{x,y\}$ in view $v$.

\subsection{Targeted Structure Completion}
While the base pixel-aligned representation successfully reconstructs deterministic surfaces, it inherently breaks down near visibility transitions where geometry is heavily occluded or truncated. To address this without incurring the prohibitive computational cost of uniform volumetric densification, we propose a lightweight targeted structure completion module that consists of several 3D completion blocks. Motivated by the efficiency of sparse operations~\cite{huang2024gaussianformer}, this module compensates for geometric incompleteness exclusively within the localized 3D uncertainty subspace $\mathcal{O}_{3d}$ derived in Section~\ref{sec:location}.

\subsubsection{Uncertainty-Aware Query Sampling}
Instead of populating the entire 3D volume with dense voxel queries or densely sampling along every individual ray, which would lead to a highly variable computational cost depending on the scene's occlusion complexity, we maintain a strict efficiency budget by extracting a fixed total number of queries globally from the 3D uncertainty subspace $\mathcal{O}_{3d}$. Specifically, we draw exactly $N_q$ discrete 3D points from the aggregated subspace to instantiate our localized Gaussian queries. The set of sampled query coordinates, denoted as $\mathcal{P}$, can be formulated as:
\begin{equation}
    \mathcal{P} = \big\{ \mathbf{p}_m \in \mathcal{O}_{3d} \;\big|\; m = 1, 2, \dots, N_q \big\}, \quad \mathbf{p}_m \sim \mathcal{U}(\mathcal{O}_{3d}),
\end{equation}
where $N_q$ is a predefined hyperparameter dictating the maximum token budget, and $\mathcal{U}(\mathcal{O}_{3d})$ denotes the uniform sampling distribution over the geometry of the lifted subspace. These sampled geometric positions $\mathbf{p}_m$ serve as the initial spatial anchors for the compensatory queries, each coupled with a learnable high-dimensional feature embedding. By decoupling the query generation from the raw pixel count and confining it to a fixed budget within the geometric ambiguity manifold, we mathematically guarantee a constant and lightweight memory footprint. These $N_q$ localized queries are subsequently optimized through $N$ targeted completion blocks.

\subsubsection{Self-context Aggregation}
To enable local geometric learning among the sampled queries, we employ 3D sparse convolution~\cite{spconv2022} for self-context aggregation. We voxelize the 3D coordinates of the sampled queries and perform 3D sparse convolutions exclusively on the occupied voxels. Since the queries are spatially restricted to the sparse $\mathcal{O}_{3d}$ subspace, this operation completely bypasses the cubic computational complexity ($\mathcal{O}(X \times Y \times Z)$) that plagues dense 3D processing pipelines. We expand the receptive field by stacking sparse convolution layers, while maintaining strict spatial sparsity to ensure maximal efficiency.

\subsubsection{Cross-view Aggregation}
To resolve the geometric ambiguity inherent in occluded regions, the localized queries must gather robust multi-view context. For each 3D Gaussian query $Q_{3d}$, we apply deformable attention (DA)~\cite{zhu2020deformable} onto the multi-view image feature maps. This cross-view aggregation effectively injects complementary visual cues from unoccluded viewpoints into the queries, disambiguating structures that are hidden in single ego-centric views.

\subsubsection{Structural Compensation}
The goal of the structural compensation layer is to decode the final attributes of the compensatory 3D Gaussians from the enriched queries. We utilize a multi-layer perceptron (MLP) to project the updated query embeddings into explicit Gaussian properties. Specifically, we refine the 3D position (mean) of each Gaussian by predicting a residual offset relative to its initial sampled location within $\mathcal{O}_{3d}$, while the other properties (opacity, scale, rotation, and color) are directly regressed from the updated query features.

By stacking the targeted completion blocks, FocusGS generates a compact, highly localized set of supplementary Gaussians dedicated solely to structural completion. Combined with the base pixel-aligned representation, this spatially decoupled approach mitigates boundary ambiguities and enforces multi-view consistency. It achieves the structural completeness of dual-branch methods while preserving the strict efficiency of pixel-based architectures.

\subsection{Training Objectives}
To render a scene, we aggregate the base pixel-aligned Gaussians $\mathcal{G}_{base}$ and the targeted compensatory Gaussians $\mathcal{G}_{comp}$ to form the full scene representation, denoted as $\mathcal{G}_{final} = \mathcal{G}_{base} \cup \mathcal{G}_{comp}$. To enable appropriate supervision specifically for the targeted Gaussians, we utilize the 2D geometric ambiguity manifold $\mathcal{O}_d$ as a spatial mask. We calculate masked photometric losses as well as a masked depth loss $L_{comp}^{dpt}$ for images and depths rendered independently from $\mathcal{G}_{comp}$. Crucially, only pixels with mask values equal to 1 are used for this targeted loss calculation, ensuring that computational resources and gradients are dedicated exclusively to ambiguity regions.

Combining this with the global photometric losses $L_{full}^{l_1}$ and $L_{full}^{lpips}$ for novel-view images rendered from our full Gaussians $\mathcal{G}_{final}$, the overall training objective $L$ can be derived as follows:
\begin{equation}
\begin{aligned}
L &= L_{full}^{l_1} + \lambda_1 L_{full}^{lpips} + \lambda_2 L_{comp}, \\
L_{comp} &= L_{comp}^{l_1} + \lambda_{c_1} L_{comp}^{lpips} + \lambda_{c_2} L_{comp}^{dpt},
\end{aligned}
\end{equation}
where $\lambda_1$ and $\lambda_2$ are weights for the LPIPS loss of $\mathcal{G}_{final}$ and the composite loss of $\mathcal{G}_{comp}$, respectively. $\lambda_{c_1}$ and $\lambda_{c_2}$ are weights for the masked LPIPS and depth losses of the targeted Gaussians $\mathcal{G}_{comp}$, respectively.

\section{Experiments}
\subsection{Experimental Settings}

\noindent\textbf{Evaluation tasks.}
We follow the protocol of Omni-Scene~\cite{wei2025omni} and evaluate FocusGS in two settings: the ego-centric setting on nuScenes~\cite{caesar2020nuscenes} and the scene-centric setting on RealEstate10K~\cite{zhou2018stereo}. For both datasets, we compare against the Gaussian-based methods~\cite{wei2025omni, charatan2024pixelsplat, chen2024mvsplat}, the light-field method AttnRend ~\cite{du2023learning}, and the NeRF-based method MuRF~\cite{xu2024murf}.

\noindent\textbf{Metrics.} 
We adopt three widely used metrics from previous 3D scene reconstruction studies~\cite{wei2025omni, chen2024mvsplat, charatan2024pixelsplat} to evaluate visual quality: peak signal-to-noise ratio (PSNR), structural similarity index (SSIM)~\cite{wang2004image}, and learned perceptual image patch similarity (LPIPS)~\cite{zhang2018unreasonable}. Higher values indicate better performance for PSNR and SSIM, whereas lower values are preferred for LPIPS. In addition, we report the Pearson correlation coefficient (PCC) to assess the geometric fidelity of reconstructed 3D scenes.

\noindent\textbf{Implementation details.} 
We implement FocusGS using the open-source Gaussian renderer~\cite{kerbl20233d}. Multi-view image features are extracted with a ResNet-50 backbone pre-trained using DINO~\cite{caron2021emerging}. For geometric ambiguity localization, depth values are clipped to the 
range $[0, 100]$, with the dilation kernel size 
and threshold as 3 and 3. We employ four targeted structure completion blocks to update within the geometric ambiguity manifold. Training is performed on two 80GB GPUs for 100k iterations with a batch size of 4 on nuScenes~\cite{caesar2020nuscenes}, and on a single 80GB GPU for 300k iterations with a batch size of 8 on RealEstate10K~\cite{zhou2018stereo}. We use AdamW~\cite{kingma2014adam} with an initial learning rate of $1\times10^{-4}$ and cosine decay. 

\begin{table*}[!t]
\centering
\footnotesize
\caption{Quantitative results of the ego-centric reconstruction task on nuScenes~\cite{caesar2020nuscenes}. PCC is reported as N/A for AttnRend~\cite{du2023learning}, since it does not produce an interpretable 3D structure for depth rendering.}
\setlength{\tabcolsep}{3.4mm}
\resizebox{0.97\textwidth}{!}{
\begin{tabular}{lccccc}
\hline
\toprule
Method & Latency(s)  & PSNR$\uparrow$ & SSIM$\uparrow$ & LPIPS$\downarrow$ & PCC$\uparrow$ \\ \hline
AttnRend~\cite{du2023learning} & {9.98}  & {20.96} & {0.533} & {0.467} & {N/A} \\
MuRF~\cite{xu2024murf} & {0.672}  & {20.34} & {0.504} & {0.433} & {-0.332} \\
pixelSplat~\cite{charatan2024pixelsplat} & {0.508}  & {21.51} & {0.616} & {0.372} & {0.001} \\
MVSplat~\cite{chen2024mvsplat} & {0.174}  & {21.61} & {0.658} & {0.295} & {0.181} \\
STORM~\cite{yang2025storm} & - & 24.56 & 0.752 & \textbf{0.217} & 0.788 \\
SCube~\cite{ren2024scube}& - & 23.85 & 0.721 & 0.258 & 0.651 \\
DrivingForward ~\cite{tian2025drivingforward} & - & 24.32 & 0.732 & 0.229 & 0.766 \\
Omni-Scene~\cite{wei2025omni} & {0.088}  & {24.27} & {0.736} & {0.237} & {0.800} \\
\rowcolor{violet!6}{FocusGS (ours)} & {\textbf{0.058}} & {\textbf{24.65}} & {\textbf{0.754}} & 0.220 & {\textbf{0.837}} \\
\bottomrule
\end{tabular}}
\label{tab:main_res}
\end{table*}

\begin{table*}[!t]
    \centering
    \caption{Quantitative results of FocusGS on RealEstate10K~\cite{zhou2018stereo} under scene-centric reconstruction setting.}
    \label{tab:rel10k_res}
    \setlength{\tabcolsep}{6pt}
    \resizebox{0.68\linewidth}{!}{%
    \begin{tabular}{lcccc}
    \toprule
    Method & PSNR$\uparrow$ & SSIM$\uparrow$ & LPIPS$\downarrow$ & PCC$\uparrow$ \\
    \midrule
    AttnRend~\cite{du2023learning}         & 24.78 & 0.820 & 0.213 & N/A   \\
    MuRF~\cite{xu2024murf}                 & 26.10 & 0.858 & 0.143 & 0.344 \\
    pixelSplat~\cite{charatan2024pixelsplat}& 25.89 & 0.858 & 0.142 & 0.285 \\
    MVSplat~\cite{chen2024mvsplat}         & 26.39 & 0.869 & 0.128 & 0.363 \\
    Omni-Scene~\cite{wei2025omni}           & 26.19 & 0.865 & 0.131 & {0.368} \\
    \rowcolor{violet!6}{FocusGS}             & 26.32 & 0.872 & 0.123 & {0.365} \\
    \rowcolor{violet!6}{MVSPlat + FocusGS}             & \textbf{27.02} & \textbf{0.892} & \textbf{0.118} & \textbf{0.374} \\
    \bottomrule
    \end{tabular}}
\end{table*}

\subsection{Main Results}
\begin{figure*}[!t]
\centering
\includegraphics[width=1.0\textwidth]{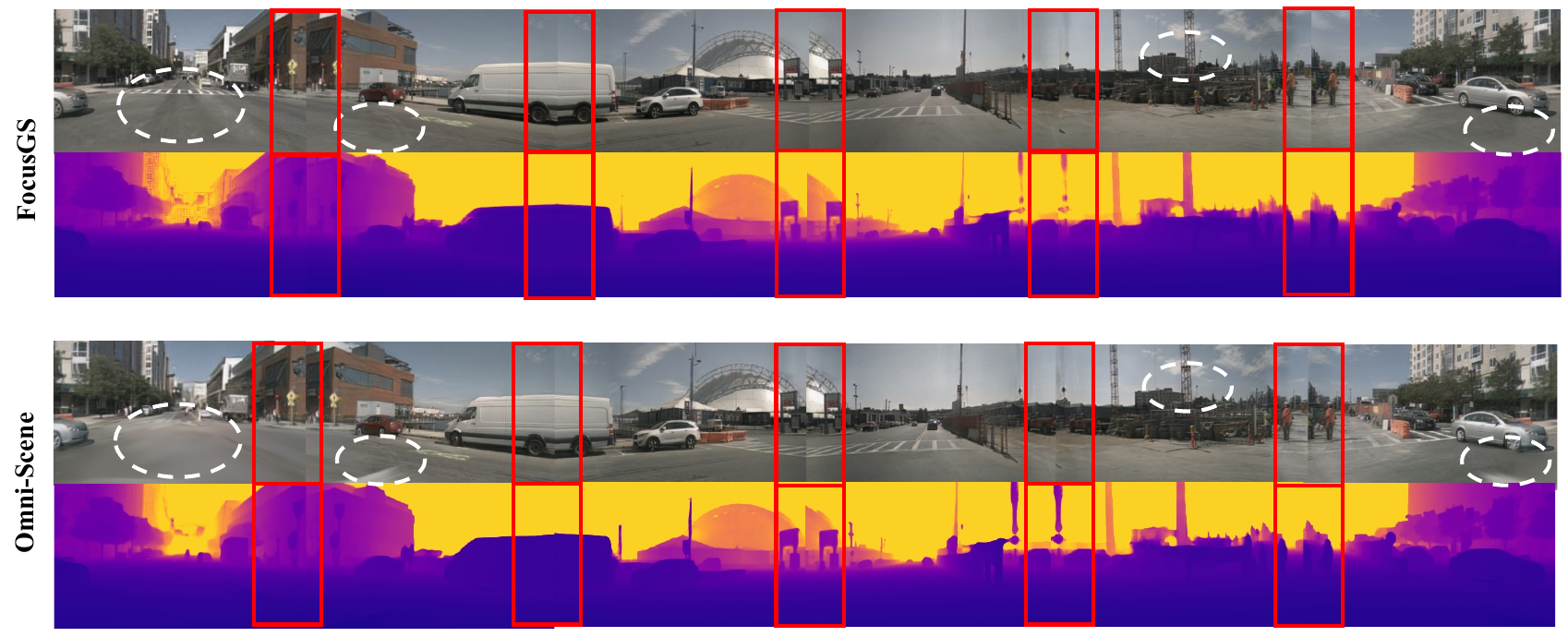}
\caption{Qualitative comparison between Omni-Scene~\cite{wei2025omni} and FocusGS. Six rendered surround views cover the full \(360^\circ\) panorama with about \(15\%\) adjacent-view overlap. Red boxes mark overlap regions, while white dashed regions indicate cases where FocusGS produces sharper and more complete reconstructions.}
\label{fig:vis}
\end{figure*}

\noindent\textbf{Results on nuScenes.}
Tab.~\ref{tab:main_res} presents a comparison between FocusGS and existing baselines on nuScenes. Compared to Omni-Scene~\cite{wei2025omni}, specifically designed for the ego-centric setting, our approach reduces rendering latency by $\sim$34\% while also achieving higher accuracy. Feed-forward 3DGS methods~\cite{chen2024mvsplat, charatan2024pixelsplat, xu2024murf, du2023learning} for sparse-view reconstruction perform worst, particularly on the PCC metric, as limited view overlap in ego-centric settings makes depth estimation unreliable. While Omni-Scene improves over MVSplat and pixelSplat, its voxel-based Gaussian branch has millions of Gaussians, even in well-observed regions where pixel-based Gaussians suffice. In contrast, our method targets refinement only to occluded regions, substantially reducing Gaussian count while preserving performance. As highlighted by the white dashed circles in Fig.~\ref{fig:vis}, Omni-Scene exhibits noticeable blurriness and structural artifacts in these specific regions, resulting in a relatively lower overall reconstruction quality. In contrast, FocusGS effectively resolves these geometric ambiguities, preserving fine details, improving geometric consistency, and achieving higher visual fidelity.

\noindent\textbf{Results on RealEstate10K.}
To further demonstrate the effectiveness and generalization of FocusGS, 
we also conduct evaluations on the RealEstate10K~\cite{zhou2018stereo} dataset, a scene-centric benchmark widely used for sparse-view reconstruction tasks. We also apply the proposed targeted structure completion to the MVSplat for further evaluation.
As shown in Tab.~\ref{tab:rel10k_res}, standalone FocusGS achieves competitive results, while \textit{MVSplat+FocusGS} achieves the best performance on all metrics. We also note that feed-forward baselines, such as pixelSplat~\cite{charatan2024pixelsplat} and MuRF~\cite{xu2024murf}, although efficient, suffer from limited geometric fidelity, particularly in terms of PCC. FocusGS underscores the effectiveness of the proposed targeted structure completion strategy compared with Omni-Scene.

\begin{table}[!t]
    \small
\caption{Ablation study of core modules under ego-centric reconstruction setting. \textit{Random S.C.} denotes random structure completion, and \textit{G.A. Localization} denotes geometric ambiguity localization.}
\centering
\label{tab:abl1}
\setlength{\tabcolsep}{6pt}
\renewcommand{\arraystretch}{1.0}
\resizebox{0.75\textwidth}{!}{%
\begin{tabular}{lccccc}
\toprule
& Method & PSNR$\uparrow$ & SSIM$\uparrow$ & LPIPS$\downarrow$ & PCC$\uparrow$ \\
\midrule
(a) & Pixel-aligned Baseline & 22.89 & 0.698 & 0.290 & 0.780   \\
(b) & (a) + Depth Init & 23.14 & 0.703 & 0.320 & 0.802 \\
(c) & (b) + Random S.C. & 23.40 & 0.708 & 0.306 & 0.810 \\
\rowcolor{violet!6} (d) & (c) + G.A. Localization & \textbf{24.65} & \textbf{0.754} & \textbf{0.220} & \textbf{0.837} \\
\bottomrule
\end{tabular}}
\end{table}

\subsection{Ablation Study}
\noindent\textbf{Overall architecture.}
The ablations on the Depth Init, Random S.C. (Structural Completion), and G.A. (Geometric Ambiguity) Localization modules are shown in Tab.~\ref{tab:abl1}. \textit{Random S.C.} brings only marginal gains over depth initialization, while G.A. Localization yields a much larger improvement, demonstrating that completion must be spatially targeted rather than randomly applied.

\begin{figure}[!t]
  \centering
  \begin{minipage}[c]{0.54\linewidth} 
    \centering
    \includegraphics[width=0.72\linewidth]{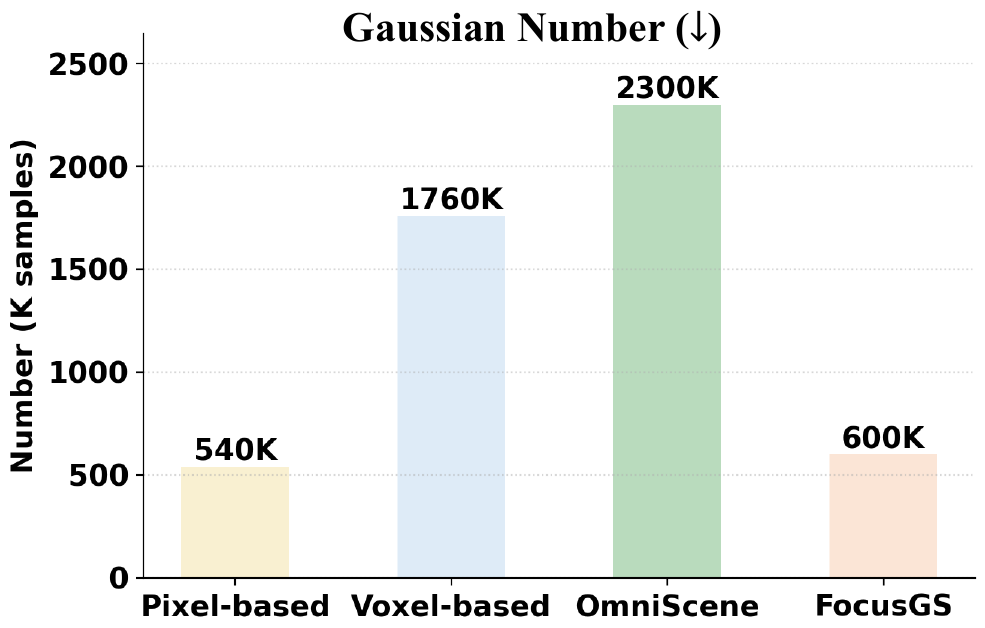}
    \captionof{figure}{Gaussian number of different methods.}
    \label{fig:num}
  \end{minipage}
  \begin{minipage}[c]{0.42\linewidth}
    \centering
    \footnotesize
    \captionof{table}{Ablation study of the blocks of targeted structure completion module under the ego-centric reconstruction setting.}
    \label{tab:abl2}
    \setlength{\tabcolsep}{1.0mm}
    \resizebox{\linewidth}{!}{
    \begin{tabular}{lcccc}
    \toprule
    Method & PSNR$\uparrow$ & SSIM$\uparrow$ & LPIPS$\downarrow$ & PCC$\uparrow$ \\
    \midrule
    w/o. S.A. & 24.04 & 0.738 & 0.234 & 0.827 \\
    w/o. C.A. & 23.30 & 0.723 & 0.265 & 0.802 \\
    \rowcolor{violet!6}{Ours} & \textbf{24.65} & \textbf{0.754} & \textbf{0.220} & \textbf{0.837} \\
    \bottomrule
    \end{tabular}
    }
  \end{minipage}
\end{figure}

\noindent\textbf{Targeted structure completion.}
We further conduct ablation studies on the components of the targeted structure completion module. We remove the self-context aggregation and cross-view aggregation layers, denoted as \textit{w/o S.A.} and \textit{w/o C.A.}, respectively. The structure compensation layer, which is responsible for decoding the updated features of Gaussians, cannot be ablated. As shown in Tab.~\ref{tab:abl2}, removing either aggregation module results in a notable performance drop. Moreover, our analysis reveals that the cross-view aggregation layer has a stronger impact on reconstruction quality compared to the self-context aggregation layer. This is because refined Gaussians iteratively obtain features both from neighboring Gaussians and from multi-view image features, and the information carried by multi-view features is substantially richer.

\begin{table}[!t]
\centering
\caption{Quantitative validation of ambiguity masks against high-error regions.}
\label{tab:abl_masks}
\setlength{\tabcolsep}{2.6mm}
\resizebox{1.0\textwidth}{!}{
\begin{tabular}{lcccccc}
\toprule
Mask Type & Recall$\uparrow$ & Precision$\uparrow$ & IoU$\uparrow$ & PSNR$\uparrow$ & SSIM$\uparrow$ & LPIPS$\downarrow$\\
\midrule
Error Mask (Oracle) & 1.00 & 1.00 & 1.00 & \textbf{27.74} & \textbf{0.812} & \textbf{0.181} \\
\midrule
pixel-aligned (FocusGS w/o TSC) & -- & -- & -- & 22.89 & 0.698 & 0.290 \\
Random Mask & 0.43 & 0.16 & 0.13 & 23.42 & 0.703 & 0.285 \\
RGB Edge Mask & 0.68 & 0.23 & 0.21 & 23.76 & 0.715 & 0.283 \\
Learned Mask & 0.55 & \textbf{0.86} & \textbf{0.50} & 24.03 & 0.732 & 0.261 \\
\rowcolor{gray!15}\textbf{Ours} & \textbf{0.82} & 0.46 & 0.42 & \textbf{24.65} & \textbf{0.754} & \textbf{0.220} \\
\bottomrule
\end{tabular}}
\end{table}

\subsection{Further Discussion}
\noindent\textbf{Effectiveness of ambiguity localization.}
We further quantitatively evaluate different ambiguity mask designs, including random sampling, RGB-edge masks, and a learned mask. For reference, we construct an oracle error mask from ground-truth error regions, which serves as an upper bound for targeted structure completion (TSC). As shown in Tab.~\ref{tab:abl_masks}, while the learned mask achieves higher precision and IoU, it tends to overlook a substantial portion of high-error structural regions, limiting its effectiveness for completion. In contrast, our depth-gradient-based mask significantly improves recall, which is more critical for identifying missing or erroneous structures. Compared with the learned mask, it increases recall by +0.27 and improves PSNR by +0.62 dB, achieving a better balance between coverage and reconstruction quality.

\noindent\textbf{Gaussian numbers.}
We conduct studies on different settings of our proposed FocusGS. When evaluating the Gaussian query number within the geometric ambiguity manifold, we observe that reconstruction quality improves as the number of queries increases, up to a critical saturation point shown in Tab.~\ref{tab:set1}. Beyond this threshold, allocating additional Gaussians yields negligible geometric gains, validating that a constrained query number acts as an optimal sweet spot for maximizing visual fidelity without compromising computational efficiency. 

\noindent\textbf{Block numbers.}
Similarly, when analyzing the network depth of the completion module, we find that a moderate number of cascaded blocks is essential for sufficient context aggregation, as shown in Tab.~\ref{tab:set2}. Utilizing a shallower network restricts representational capacity, whereas excessively deep architectures lead to a degradation in reconstruction quality, likely due to optimization difficulties or overfitting within the sparse feature space. Consequently, our final framework adopts these balanced configurations to ensure robust structural compensation while maintaining strict efficiency.

\begin{table}[!t]
  \centering
  \begin{minipage}[c]{0.48\linewidth} 
    \centering
    \footnotesize
    \caption{Impact of the number of Gaussians in targeted structure completion.}
    \label{tab:set1}
    \setlength{\tabcolsep}{1.5mm}
    \resizebox{0.9\linewidth}{!}{
    \begin{tabular}{ccccc}
        \toprule
        Nums \quad & PSNR$\uparrow$ & SSIM$\uparrow$ & LPIPS$\downarrow$ & PCC$\uparrow$ \\
        \hline
        10k  \quad & 23.79  & 0.747 & 0.231 & 0.830 \\
        60k \quad & 24.65 & 0.754 & 0.220 & 0.837  \\
        \rowcolor{violet!6}{120k}
        \quad & 24.67 & 0.756 & 0.218 & 0.842
        \\
        \bottomrule
        \end{tabular}
    }
  \end{minipage}
  \begin{minipage}[c]{0.48\linewidth}
    \centering
    \footnotesize
    \caption{Impact of the number of completion blocks.}
    \label{tab:set2}
    \setlength{\tabcolsep}{1.5mm}
    \resizebox{0.9\linewidth}{!}{
    \begin{tabular}{ccccc}
        \toprule
        Nums \quad & PSNR$\uparrow$ & SSIM$\uparrow$ & LPIPS$\downarrow$ & PCC$\uparrow$ \\
        \hline
        1  \quad & 23.76  & 0.747 & 0.224 & 0.828 \\
        \rowcolor{violet!6}{4}  \quad & \textbf{24.65} & \textbf{0.754} & \textbf{0.220} & \textbf{0.837} \\
        6  \quad & 24.56  & 0.744 & 0.226 & 0.841 \\
        \bottomrule
        \end{tabular}
    }
  \end{minipage}
\end{table}

\begin{figure}[!t]
  \centering
  \begin{minipage}[c]{0.5\linewidth} 
    \centering
    \includegraphics[width=0.85\linewidth]{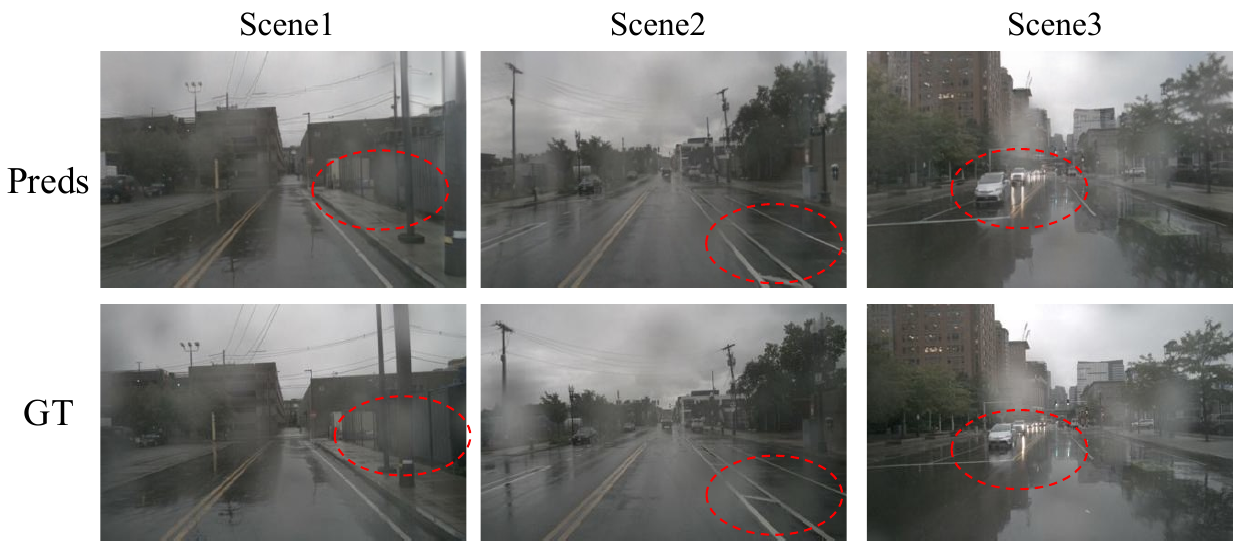}
    \captionof{figure}{Failure cases of our FocusGS.}
    \label{fig:failure}
  \end{minipage}
  \begin{minipage}[c]{0.48\linewidth}
    \centering
    \footnotesize
    \captionof{table}{Impact of kernel size of 2D geometric ambiguity manifold.}
    \label{tab:kernel1}
    \setlength{\tabcolsep}{1.0mm}
    \resizebox{0.85\linewidth}{!}{
        \begin{tabular}{ccccc}
        \toprule
        Size \quad & PSNR$\uparrow$ & SSIM$\uparrow$ & LPIPS$\downarrow$ & PCC$\uparrow$ \\
        \hline
        1  \quad & 23.78  & 0.736 & 0.232 & 0.835 \\
        \rowcolor{violet!6}{3}  \quad & \textbf{24.65} & \textbf{0.754} & \textbf{0.220} & \textbf{0.837} \\
        7 \quad & 24.13  & 0.750 & 0.226 & 0.832 \\
        \bottomrule
        \end{tabular}
    }
  \end{minipage}
\end{figure}

\noindent\textbf{Kernel size.}
As shown in Tab.~\ref{tab:kernel1}, we evaluate the morphological dilation kernel size defining the 2D geometric ambiguity manifold. A minimal kernel yields suboptimal quality by missing the broader uncertainty bands surrounding severe occlusions. Conversely, an excessively large kernel over-expands the manifold into flat surfaces. This over-expansion dilutes computational efficiency and degrades performance by introducing optimization interference into well-reconstructed regions. Thus, a moderate kernel size is optimal, accurately encapsulating true ambiguities to maximize visual fidelity without adding structural noise.

\noindent\textbf{Failure cases.}
While robust in standard driving scenarios, FocusGS struggles under adverse weather conditions like rain, shown in Fig.~\ref{fig:failure}. Water droplets on camera lenses introduce severe optical blur and noise directly into the input source images. This sensor-level corruption disrupts multi-view consistency, making it difficult for the network to disentangle 2D lens artifacts from the true 3D underlying geometry. Consequently, as highlighted by the red dashed circles, FocusGS erroneously bakes this input-induced ambiguity into the 3D representation, leading to blurry and distorted reconstructions in the affected regions. Addressing such weather-induced sensor noise through explicit artifact modeling remains an important direction for future work.

\section{Conclusion}
In this paper, we presented FocusGS, a highly efficient framework for sparse-view 3D reconstruction in ego-centric autonomous driving that resolves the computational redundancy inherent in uniform volumetric processing. To bridge the gap between pixel-based efficiency and volumetric structural completeness, we introduced a novel targeted structure completion paradigm. By formally defining a geometric ambiguity manifold to accurately localize high-uncertainty regions, FocusGS decouples structural completion from deterministic surfaces. A lightweight targeted structure completion module then optimizes continuous queries strictly within this sparse 3D topological subspace, circumventing the massive overhead of dense voxel grids. Extensive experiments demonstrate that FocusGS establishes a superior efficiency-quality trade-off. Further ablation studies validate that targeted structure completion is more effective than random completion.
We hope this work can inspire future research on more adaptive and spatiotemporal Gaussian completion for complex dynamic driving scenes.

\noindent\textbf{Limitations:} Despite its efficiency, FocusGS still relies on the quality of the estimated geometric ambiguity manifold, which may become less reliable under highly dynamic objects, extreme lighting changes, or corrupted visual inputs. In particular, adverse weather conditions such as rain can introduce lens artifacts and severe blur, causing the model to confuse sensor-level noise with true 3D geometry; extending the current spatial completion strategy to a spatiotemporal and artifact-aware framework remains an important direction for future work.

\section*{Acknowledgements}
This work was supported by the National Natural Science Foundation of China (Grant Nos. 62322113, 62376156), as well as the Shanghai Municipal Special Program for Basic Research on General AI Foundation Models (Grant No. 2025SHZDZX025G15).

\bibliographystyle{splncs04}
\bibliography{references}
\end{document}